\documentclass[conference]{IEEEtran}

\usepackage{cite}
\usepackage{amsmath,amssymb,amsfonts}
\usepackage{textcomp}
\usepackage{booktabs}
\usepackage{array}
\usepackage{algorithm}
\usepackage{algpseudocode}
\usepackage{graphicx}
\usepackage[hidelinks]{hyperref}

\newcolumntype{L}[1]{>{\raggedright\arraybackslash}p{#1}}

\begin{document}

\title{CLOCs-R: Radar-Assisted Camera-LiDAR Fusion for Long-Range Object Detection}

\author{\IEEEauthorblockN{Richard Beuck}
\IEEEauthorblockA{\textit{U.A. Whitaker College of Engineering} \\
\textit{Florida Gulf Coast University}\\
Fort Myers, FL, USA \\
rjbeuck1394@eagle.fgcu.edu}
\and
\IEEEauthorblockN{John Nguyen}
\IEEEauthorblockA{\textit{U.A. Whitaker College of Engineering} \\
\textit{Florida Gulf Coast University}\\
Fort Myers, FL, USA \\
jdnguyen1381@eagle.fgcu.edu}
\and
\IEEEauthorblockN{Sudip Dhakal}
\IEEEauthorblockA{\textit{U.A. Whitaker College of Engineering} \\
\textit{Florida Gulf Coast University}\\
Fort Myers, FL, USA \\
sdhakal@fgcu.edu}
}

\maketitle

\section{Related Work}\label{sec:related}

This section reviews the camera-LiDAR and radar fusion methods most relevant to our proposed approach, grouped by the sensors used and the stage at which they are fused. Table~\ref{tab:compare} summarizes the method, inputs, dataset, key result, and main limitation of each work referenced in this section.

\begin{table*}[t]
\caption{Comparison of Reviewed Work (C = Camera, L = LiDAR, R = Radar, nuSc.\ = nuScenes, VoD = View-of-Delft)}
\label{tab:compare}
\centering
\scriptsize
\setlength{\tabcolsep}{4pt}
\begin{tabular}{@{}llllll@{}}
\toprule
\textbf{Work} & \textbf{Method} & \textbf{Input} & \textbf{Dataset} & \textbf{Key Result} & \textbf{Limitation} \\
\midrule
SECOND~\cite{yan2018second} & Sparse 3D convolution on voxels & L & KITTI & 73.66 mod.\ car 3D AP at 20 FPS & Misses cars with $<$10 points \\
PointPainting~\cite{vora2020pointpainting} & Image segmentation scores on points & C+L & KITTI, nuSc. & mAP 40.1 $\rightarrow$ 46.4 (nuSc.\ test) & Needs LiDAR points on the object \\
CLOCs~\cite{pang2020clocs} & Rescores pre-NMS 2D--3D pairs & C+L & KITTI & +5.90 mod.\ car 3D AP; best at 40--50\,m & Cannot add missed 3D boxes \\
Fast-CLOCs~\cite{pang2022fastclocs} & Light 2D detector cued by 3D boxes & C+L & KITTI, nuSc. & mAP 58.0 $\rightarrow$ 63.1; 8--13\,Hz & Needs high 3D recall \\
BEVFusion~\cite{liu2023bevfusion} & Shared BEV feature fusion & C+L & nuSc., Waymo & 70.2 mAP, 72.9 NDS (test) & Depth errors misalign features \\
\midrule
CRF-Net~\cite{nobis2019crfnet} & Radar image channels, multi-level & C+R & nuSc., TUM & mAP 43.47 $\rightarrow$ 43.95 & 2D only; small gain \\
RRPN~\cite{nabati2019rrpn} & Distance-scaled radar anchors & C+R & nuSc. & Beats Selective Search; $>$100$\times$ faster & 2D only; needs a radar return \\
CenterFusion~\cite{nabati2021centerfusion} & Frustum radar--center association & C+R & nuSc. & NDS 40.0 $\rightarrow$ 44.9 (test) & Only marginal mAP gain \\
RADIANT~\cite{long2023radiant} & Radar-to-center offsets, depth fusion & C+R & nuSc. & PGD mAP 36.0 $\rightarrow$ 38.0 (test) & Refines position only \\
CRN~\cite{kim2023crn} & Radar-guided BEV transform & C+R & nuSc. & 62.4 NDS; beats LiDAR at 30--100\,m$^{\dagger}$ & LiDAR depth used in training \\
CR3DT~\cite{baumann2024cr3dt} & Radar pillars + BEV features, tracking & C+R & nuSc. & +5.3 mAP, +14.9 AMOTA (val) & 21.3 mAP below LiDAR \\
RCM-Fusion~\cite{kim2024rcmfusion} & BEV + instance-level radar refinement & C+R & nuSc. & 58.7 NDS, 50.6 mAP (test) & 2.3 FPS; 87.6\,M parameters \\
\midrule
RadarNet~\cite{yang2020radarnet} & Early and attention-based late fusion & L+R & nuSc., DenseRadar & Car AP 82.3 $\rightarrow$ 84.5 (val) & Only +0.8 AP at 70--100\,m \\
FUTR3D~\cite{chen2023futr3d} & Sensor-agnostic query sampler & Any & nuSc. & C+R 39.9 vs.\ 33.2 mAP (val) & No C+L+R results \\
\midrule
View-of-Delft~\cite{palffy2022vod} & 4D radar dataset & C+L+R & VoD & Radar mAP 38.0 $\rightarrow$ 29.1 without Doppler & Labels only to 50\,m \\
K-Radar~\cite{paek2022kradar} & 4D radar tensor dataset & C+L+R & K-Radar & Heavy snow AP: radar 44.5 vs.\ LiDAR 29.7 & Labels to 120\,m; $\sim$12\,TB \\
aiMotive~\cite{matuszka2023aimotive} & Long-range dataset with baselines & C+L+R & aiMotive & Night AP $>$75\,m: 0.464 $\rightarrow$ 0.550 & Simple fusion; worse in rain \\
\bottomrule
\multicolumn{6}{@{}l}{$^{\dagger}$CRN outperforms LiDAR-only CenterPoint on cars at 30--100\,m, in the authors' extended 100\,m evaluation of nuScenes.}
\end{tabular}
\end{table*}

\subsection{Camera-LiDAR Fusion}
Most camera-LiDAR fusion methods build on a LiDAR detector. SECOND~\cite{yan2018second} replaced the dense 3D convolutions of earlier voxel networks with sparse convolutions, reaching 73.66 moderate difficulty car 3D AP on the KITTI test set~\cite{geiger2012kitti} at about 20 frames per second. Its authors note, however, that it misses many distant cars that return fewer than 10 LiDAR points~\cite{yan2018second}.

Fusion methods differ mainly in where the two modalities meet. At the point level, PointPainting~\cite{vora2020pointpainting} projects each LiDAR point into an image segmentation map and appends the class scores to the point, so that any LiDAR detector can use them. This raised PointPillars+ from 40.1 to 46.4 mAP on the nuScenes test set~\cite{caesar2020nuscenes,vora2020pointpainting}, but painting can only help objects that already have LiDAR points. At the feature level, BEVFusion~\cite{liu2023bevfusion} lifts camera features into a bird's-eye-view (BEV) grid shared with LiDAR and makes this transform about 40 times faster, reaching 70.2 mAP and 72.9 NDS on nuScenes test, although inaccurate camera depth leaves local misalignment between the two feature maps.

At the decision level, CLOCs~\cite{pang2020clocs} takes the candidates of a pretrained 2D detector and a pretrained 3D detector before non-maximum suppression (NMS). Each overlapping pair is encoded by its image-plane IoU, both confidence scores, and the object's distance, and a small network learns to rescore the 3D candidates. Fusing SECOND with Cascade R-CNN raised moderate car 3D AP on the KITTI test set by 5.90 points, with the largest gains at 40-50\,m~\cite{pang2020clocs}. Because late fusion only prunes and rescores, however, CLOCs cannot recover objects that the 3D detector never proposed~\cite{pang2020clocs}. Fast-CLOCs~\cite{pang2022fastclocs} replaces the separate 2D detector with a lightweight one that uses projected 3D candidates as proposals, running at 8-13\,Hz on a desktop GPU and raising CenterPoint from 58.0 to 63.1 mAP on nuScenes test. This design makes the 2D branch depend even more on the recall of the 3D detector~\cite{pang2022fastclocs}.

\subsection{Camera-Radar Fusion}
Radar complements both cameras and LiDAR. Automotive radars detect targets at ranges of 250\,m or more, measure radial velocity directly through the Doppler effect, and are less affected by poor lighting and weather~\cite{caesar2020nuscenes,yao2024radarcamera}. Their returns, however, are sparse, noisy, and lack reliable height information~\cite{nabati2021centerfusion,yang2020radarnet}.

Early work used radar to guide 2D detection. CRF-Net~\cite{nobis2019crfnet} projects radar returns into the image as additional input channels and lets the network learn at which feature level to fuse them, while RRPN~\cite{nabati2019rrpn} converts each projected radar return into distance-scaled anchor boxes, generating proposals far faster than Selective Search. Both produce only 2D boxes, and CRF-Net gained little from unfiltered radar~\cite{nobis2019crfnet,nabati2019rrpn}.

Object-level methods attach radar to camera detections. CenterFusion~\cite{nabati2021centerfusion} associates radar returns with camera-detected object centers through a frustum and uses them to re-estimate depth and velocity. This improved NDS and sharply reduced velocity error on nuScenes, but mAP improved only marginally, since radar can only refine objects the camera has already found. RADIANT~\cite{long2023radiant} instead predicts the offset from each radar return to the object center and fuses radar depth into a frozen monocular detector, although only object position is refined.

BEV-level methods fuse radar earlier so that it can help locate objects. CRN~\cite{kim2023crn} uses radar occupancy to guide the camera-to-BEV transform and fuses the two maps with deformable cross-attention. For cars at 30-100\,m it even outperforms the LiDAR detector CenterPoint, although its training still relies on depth supervision from projected LiDAR points~\cite{kim2023crn}. CR3DT~\cite{baumann2024cr3dt} concatenates radar pillars with camera BEV features and passes the improved velocity estimates to a tracker, improving both detection and tracking with little added latency, but it still trails CenterPoint by a wide margin. RCM-Fusion~\cite{kim2024rcmfusion} adds an instance-level stage that refines each proposal using nearby radar points, which improves accuracy at the cost of a large and slow model.

\newpage
\subsection{Radar-LiDAR Fusion}
Radar has also been fused directly with LiDAR. RadarNet~\cite{yang2020radarnet} combines the two sensors through early voxel-level fusion and late attention-based fusion, improving car AP on nuScenes over the best LiDAR-only model while lowering velocity error. Its authors observe that LiDAR point density decreases with range, whereas radar provides more observations at long range, yet on their 100\,m DenseRadar dataset radar added little at 70-100\,m~\cite{yang2020radarnet}. FUTR3D~\cite{chen2023futr3d} uses a modality-agnostic query sampler so that a single transformer decoder can use any combination of cameras, LiDAR, and radar. Its camera-radar model outperforms CenterFusion on nuScenes, but the paper reports no results using all three sensors.

\subsection{Datasets with Radar and Long-Range Evaluation}
Long-range gains are also difficult to measure. KITTI has no radar~\cite{geiger2012kitti}, and in nuScenes the LiDAR is rated only to 70\,m while each radar reaches 250\,m~\cite{caesar2020nuscenes}. Two recent datasets provide 4D radar, which measures elevation in addition to range, azimuth, and Doppler velocity. View-of-Delft~\cite{palffy2022vod} pairs a 3+1D radar with a 64-beam LiDAR and a stereo camera in dense urban traffic. A radar-only detector trained on it still trails its LiDAR counterpart, and removing Doppler velocity from the radar input markedly lowers accuracy, but annotations extend only to 50\,m~\cite{palffy2022vod}. K-Radar~\cite{paek2022kradar} records the full 4D radar tensor together with LiDAR and cameras in fog, rain, and snow, with labels to 120\,m, and its radar-only baseline outperforms a LiDAR detector in heavy snow. The aiMotive dataset~\cite{matuszka2023aimotive} was designed specifically for long-range perception, with four cameras, a 64-beam LiDAR, two long-range radars, and annotations out to about 200\,m across highway, urban, night, and rain scenes. In its baselines, adding radar to LiDAR barely changed overall AP beyond 75\,m but raised it substantially at night, while it reduced accuracy in heavy rain, which the authors attribute to their simple fusion method~\cite{matuszka2023aimotive}.

\subsection{Research Gap}
Radar helps most where cameras and LiDAR are weakest, at long range and at night~\cite{kim2023crn,matuszka2023aimotive}, yet long-range accuracy remains low: even CRN reaches only 7.0 car AP at 60-100\,m on nuScenes~\cite{kim2023crn}. Most radar fusion methods pair radar with cameras alone and train the full network end to end~\cite{kim2023crn,baumann2024cr3dt,kim2024rcmfusion}, while the few that use all three sensors either report no three-sensor results~\cite{chen2023futr3d} or rely on simple fusion~\cite{matuszka2023aimotive}. Public datasets also limit long-range study: the 4D radar datasets View-of-Delft and K-Radar label objects only to 50 and 120\,m~\cite{palffy2022vod,paek2022kradar}, so among the datasets reviewed here, only aiMotive combines camera, LiDAR, and long-range radar with annotations out to 200\,m~\cite{matuszka2023aimotive}. Candidate-level fusion such as CLOCs~\cite{pang2020clocs,pang2022fastclocs} is lightweight, works with any pretrained detectors, and gains the most at long range, but it has been studied only with cameras and LiDAR, and its long-range benefit has been analyzed only out to 50\,m~\cite{pang2020clocs}. To our knowledge, no prior work adds radar evidence to candidate-level fusion. CLOCs-R plans to address this gap by adding radar-derived features, namely the radar returns supporting each 3D candidate and their Doppler consistency, to the CLOCs pair encoding while keeping all detectors frozen, and by evaluating detection accuracy by distance out to 200\,m on the aiMotive dataset.

\section{Methodology}\label{sec:method}
CLOCs-R extends the candidate-level fusion of CLOCs~\cite{pang2020clocs} with radar evidence. A pretrained 2D image detector and a pretrained 3D LiDAR detector produce candidates before non-maximum suppression (NMS), radar returns are associated with each 3D candidate, and a small fusion network rescores the 3D candidates using geometric, semantic, and radar cues. Both detectors remain frozen, so only the fusion network is trained. Because it is not known in advance which radar cues are most informative, CLOCs-R defines a set of candidate radar features and evaluates several combinations of them as inputs to the fusion network. This section describes candidate generation, the radar features, the fusion network and its training, and the evaluation protocol.

\subsection{Overview}
\begin{figure*}[t]
\centering
\includegraphics[width=\textwidth]{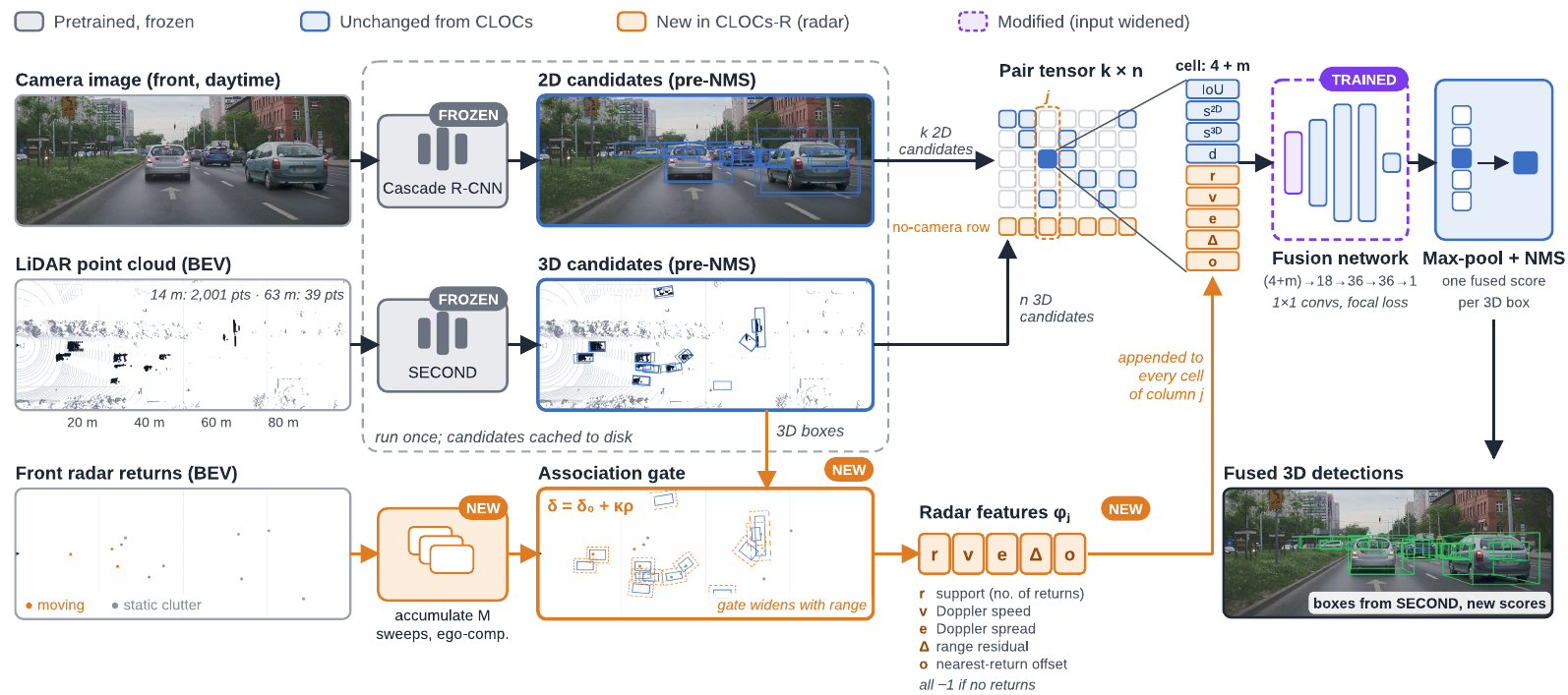}
\caption{CLOCs-R architecture, shown on a frame from the aiMotive dataset~\cite{matuszka2023aimotive}. The frozen camera and LiDAR detectors produce candidates before NMS, radar returns gated around each 3D candidate give its radar features $\boldsymbol{\phi}_j$, and these are appended to every cell of column $j$ of the pair tensor before the fusion network rescores each 3D box. Candidate boxes are generated from ground-truth labels for illustration.}
\label{fig:arch}
\end{figure*}

For each frame, the inputs are a front camera image, a LiDAR sweep, and the radar returns from the most recent radar sweeps, together with the calibration between the sensors. The 2D detector outputs $k$ axis-aligned image boxes with scores $s^{2D}_i$, and the 3D detector outputs $n$ oriented boxes with scores $s^{3D}_j$. Following CLOCs~\cite{pang2020clocs}, every 2D-3D pair is encoded as one cell of a sparse $k \times n$ tensor, a stack of $1{\times}1$ convolutions maps each cell to a logit, and max-pooling over the 2D axis yields one fused score per 3D candidate. CLOCs-R appends radar channels to each cell, drawn from the candidate features in Section~\ref{sec:radarfeat}. Because these channels describe the 3D candidate rather than the pair, they are also available to candidates that have no overlapping 2D detection, which allows radar to support distant objects that the camera detector misses. The fused scores replace the original 3D scores, and standard NMS produces the final detections. As in CLOCs, box geometry is inherited from the 3D detector and only confidence scores change. Fig.~\ref{fig:arch} shows the pipeline on a clear daytime urban frame from the aiMotive dataset, using the front camera, the LiDAR, and the front long-range radar, with the bird's-eye views drawn out to 100\,m. LiDAR density falls quickly with range: a car at 14\,m receives about 2,000 LiDAR points, whereas a car at 63\,m receives 39. The radar returns in the same frame extend to 171\,m, beyond the region shown. Radar returns are separated into moving objects and static clutter by their ego-compensated Doppler velocity and are matched to each 3D candidate through a gate that widens with range. The resulting radar features, namely the support $r$, Doppler speed $v$, Doppler spread $e$, range residual $\Delta$, and nearest-return offset $o$ defined in Section~\ref{sec:radarfeat}, are appended to every cell of column $j$, including the no-camera row. Components that CLOCs-R adds are shown in orange, and the fusion network, whose input is widened, in purple. The aiMotive data are used under the CC BY-NC-SA 4.0 license.

\subsection{Candidate Generation}
\textit{3D detector.} We use SECOND~\cite{yan2018second} as the 3D detector, the same detector used in the CLOCs configuration that we reproduce on KITTI. For aiMotive~\cite{matuszka2023aimotive}, the detector is retrained with its point-cloud range extended to cover the dataset's annotation range and its anchor sizes set from the aiMotive training labels. 
All anchor-level predictions are exported before NMS with no score threshold, so that the fusion stage sees every candidate the detector produces.

\textit{2D detector.} We use Cascade R-CNN~\cite{cai2018cascade} as the 2D detector, implemented in MMDetection~\cite{chen2019mmdetection}. Because aiMotive provides only 3D annotations, 2D training labels are generated by projecting each annotated 3D box into the front camera image with the dataset's calibration and taking the tightest enclosing axis-aligned rectangle. Boxes whose projections fall outside the image are discarded. At inference, the detector's outputs are likewise taken before NMS and with a very low score threshold, and the top $K_{2D} = 200$ candidates per image are kept, matching the CLOCs implementation.

\textit{Caching.} Both detectors are run once over every frame, and their candidates are stored on disk. The fusion network is then trained and evaluated from these cached candidates, which keeps its training independent of the cost of the detectors.

\subsection{Radar Evidence Features}\label{sec:radarfeat}
\textit{Preprocessing.} Radar returns are transformed into the LiDAR coordinate frame using the extrinsic calibration. Because a single radar sweep is sparse, returns from the last $M$ sweeps are accumulated and transformed into the current frame using the ego-vehicle poses. 
Each return $q$ provides a bird's-eye-view position $\mathbf{p}_q$ and a measured radial velocity $v^{m}_q$. The aiMotive radars report the range, azimuth, elevation, radial speed (in m/s), and reflectivity of every return~\cite{matuszka2023aimotive}, so a Doppler measurement is available for each return that CLOCs-R uses. This measurement is important: on View-of-Delft, removing Doppler velocity from the input of a radar-only detector lowered mAP from 38.0 to 29.1~\cite{palffy2022vod}. The reported speed is relative to the ego vehicle: in the aiMotive sample sequence, returns from static structures read about $-30$\,m/s at highway speed, close to the negative ego speed, while returns on annotated vehicles match their labeled relative velocities. The radial velocity is therefore compensated as
\begin{equation}
\tilde{v}_q = v^{m}_q + \mathbf{u}^{\top} \frac{\mathbf{p}_q}{\lVert \mathbf{p}_q \rVert},
\label{eq:comp}
\end{equation}
where $\mathbf{u}$ is the ego velocity in the BEV plane, obtained by differencing the consecutive GNSS-INS ego poses provided with the dataset, so that static objects have $\tilde{v}_q \approx 0$. We verify the compensation on each sequence by checking that returns from static structures, such as guardrails and signs, have near-zero $\tilde{v}_q$.

\textit{Association.} For each 3D candidate $j$, we form a gate $\mathcal{G}_j$ by enlarging its BEV footprint by a margin
\begin{equation}
\delta_j = \delta_0 + \kappa \rho_j,
\label{eq:gate}
\end{equation}
where $\rho_j$ is the candidate's distance from the ego vehicle. The range-dependent term accounts for radar's lateral position error, which grows with distance for a fixed angular resolution. The parameters $\delta_0$ and $\kappa$ are selected on a held-out subset of the training scenes. Let $\mathcal{Q}_j$ be the set of radar returns inside $\mathcal{G}_j$, with $N_j = |\mathcal{Q}_j|$.

\textit{Candidate features.} We define five candidate features that summarize the radar evidence for candidate $j$, each computed only from the positions and compensated radial velocities of the returns in $\mathcal{Q}_j$:
\begin{itemize}
\item \emph{Support}, $r_j = \log(1 + N_j)$, which measures how many returns fall in or near the box. The logarithm limits the influence of large vehicles that produce many returns.
\item \emph{Doppler speed}, $v_j = \operatorname{median}_{q \in \mathcal{Q}_j} \lvert \tilde{v}_q \rvert / v_{\max}$, which helps distinguish moving objects from static clutter such as guardrails and signs.
\item \emph{Doppler spread}, $e_j = \operatorname{median}_{q \in \mathcal{Q}_j} \lvert \tilde{v}_q - \bar{v}_j \rvert / v_{\max}$, where $\bar{v}_j$ is the median of $\tilde{v}_q$. Returns from a single rigid object share one velocity, so a small spread indicates that the returns are consistent with one object rather than a mix of neighboring objects and clutter.
\item \emph{Range residual}, $\Delta_j = \lvert \operatorname{median}_{q \in \mathcal{Q}_j} \lVert \mathbf{p}_q \rVert - \rho_j \rvert / \delta_j$, which compares radar's direct range measurement with the range of the 3D box and can indicate a candidate placed at the wrong depth.
\item \emph{Nearest-return offset}, $o_j = \min_{q \in \mathcal{Q}_j} \lVert \mathbf{p}_q - \mathbf{c}_j \rVert / \delta_j$, where $\mathbf{c}_j$ is the box center in BEV, which measures how closely the radar evidence agrees with the box location.
\end{itemize}
The medians reduce the effect of outlying returns, and dividing by the constant $v_{\max}$ or the gate margin $\delta_j$ keeps every feature on a comparable scale. When $N_j = 0$, we set $r_j = 0$ and all other radar features to $-1$, following the CLOCs convention of using $-1$ to mark missing evidence so that it is distinguishable from a measured zero.

\textit{Feature combinations.} Table~\ref{tab:featsets} lists the radar feature sets we evaluate. They range from single features, which show what each cue contributes on its own, to the support and Doppler pair proposed for CLOCs-R and the full set of five. Each set is trained and evaluated separately with an otherwise identical network and training schedule. The set used for the final CLOCs-R model is selected by its center-distance AP beyond 50\,m on held-out training scenes, and all sets are reported on the validation split.

\begin{table}[t]
\caption{Radar Feature Sets Evaluated}
\label{tab:featsets}
\centering
\footnotesize
\setlength{\tabcolsep}{5pt}
\begin{tabular}{@{}lcccccc@{}}
\toprule
\textbf{Set} & $\boldsymbol{r_j}$ & $\boldsymbol{v_j}$ & $\boldsymbol{e_j}$ & $\boldsymbol{\Delta_j}$ & $\boldsymbol{o_j}$ & \textbf{Channels} \\
\midrule
R0 (CLOCs, no radar) & & & & & & 4 \\
R1 & \checkmark & & & & & 5 \\
R2 & & \checkmark & & & & 5 \\
R3 & \checkmark & \checkmark & & & & 6 \\
R4 & \checkmark & \checkmark & \checkmark & & & 7 \\
R5 & \checkmark & & & \checkmark & \checkmark & 7 \\
R6 & \checkmark & \checkmark & & \checkmark & & 7 \\
R7 (all) & \checkmark & \checkmark & \checkmark & \checkmark & \checkmark & 9 \\
\bottomrule
\end{tabular}
\end{table}

\subsection{Fusion Network and Training}
Each non-empty cell of the input tensor holds
\begin{equation}
\mathbf{T}_{ij} = \left[\, \mathrm{IoU}_{ij},\ s^{2D}_i,\ s^{3D}_j,\ d_j,\ \boldsymbol{\phi}_j^{\top} \,\right],
\label{eq:tensor}
\end{equation}
where $\mathrm{IoU}_{ij}$ is the overlap between 2D box $i$ and the image projection of 3D box $j$, $d_j$ is the normalized BEV distance of candidate $j$ from the LiDAR, and $\boldsymbol{\phi}_j \in \mathbb{R}^{m}$ holds the $m$ radar features of the feature set being evaluated. As in CLOCs, cells with zero IoU are discarded, and each 3D candidate without an overlapping 2D box is kept in a reserved row with $\mathrm{IoU}$ and $s^{2D}$ set to $-1$~\cite{pang2020clocs}. The radar channels are filled in every cell of column $j$, including this reserved row. The fusion network keeps the CLOCs architecture of four $1{\times}1$ convolutions with ReLU activations ($(4{+}m){\to}18{\to}36{\to}36{\to}1$), so only the input width changes with the feature set. The network therefore has between 2,161 and 2,233 parameters across the sets in Table~\ref{tab:featsets}, compared with 2,143 for CLOCs. The fused score of candidate $j$ is
\begin{equation}
\hat{s}_j = \max_{i} f_{\theta}(\mathbf{T}_{ij}),
\label{eq:fused}
\end{equation}
where $f_{\theta}$ is the fusion network and the maximum is taken over all non-empty cells in column $j$. Algorithm~\ref{alg:clocsr} summarizes the per-frame procedure.

\begin{algorithm}[t]
\caption{CLOCs-R inference for one frame}
\label{alg:clocsr}
\begin{algorithmic}[1]
\Require image $I$, LiDAR sweep $P$, last $M$ radar sweeps, ego poses, radar feature set $\mathcal{S}$
\State $\{(B^{2D}_i, s^{2D}_i)\} \gets$ 2D detector$(I)$ \Comment{pre-NMS, top $K_{2D}$}
\State $\{(B^{3D}_j, s^{3D}_j)\} \gets$ 3D detector$(P)$ \Comment{pre-NMS}
\State $\mathcal{Q} \gets$ accumulate radar sweeps; compensate Doppler \eqref{eq:comp}
\For{each 3D candidate $j$}
  \State $\delta_j \gets \delta_0 + \kappa \rho_j$ \eqref{eq:gate}
  \State $\mathcal{Q}_j \gets \{q \in \mathcal{Q} : \mathbf{p}_q \in \mathcal{G}_j\}$
  \State $\boldsymbol{\phi}_j \gets$ features in $\mathcal{S}$ from $\mathcal{Q}_j$ \Comment{$-1$ if $\mathcal{Q}_j = \emptyset$}
\EndFor
\State $\mathbf{T} \gets$ cells $\mathbf{T}_{ij}$ with $\mathrm{IoU}_{ij} > 0$, plus no-camera row \eqref{eq:tensor}
\State $\hat{s}_j \gets \max_i f_{\theta}(\mathbf{T}_{ij})$ for all $j$ \eqref{eq:fused}
\State \Return NMS$\big(\{(B^{3D}_j, \hat{s}_j)\}\big)$
\end{algorithmic}
\end{algorithm}

Training targets follow the CLOCs implementation: a 3D candidate is positive if its BEV IoU with a ground-truth box is at least 0.7 and negative if it is at most 0.5, and candidates in between are ignored. Because nearly all candidates are background, the network is trained with focal loss~\cite{lin2017focal} with $\alpha = 0.25$ and $\gamma = 2$. We use the Adam optimizer~\cite{kingma2015adam} with a one-cycle learning-rate schedule (maximum learning rate $3{\times}10^{-3}$), weight decay of 0.01, and a batch size of one frame.

\subsection{Datasets and Evaluation Protocol}
\textit{KITTI verification.} Before modifying the method, we reproduce camera-LiDAR CLOCs on KITTI~\cite{geiger2012kitti} with the official implementation, confirming that our software environment matches the published results. This step uses the standard split of 3,712 training and 3,769 validation frames.

\textit{aiMotive.} All CLOCs-R experiments use the aiMotive dataset~\cite{matuszka2023aimotive}, which provides synchronized camera, 64-beam LiDAR, and long-range radar data, GNSS-INS ego poses, and 3D annotations out to about 200\,m. We use the front camera, the LiDAR, and the front long-range radar. We use its official split of 151 training and 25 validation scenes (21,402 and 5,181 annotated frames), hold out a subset of training scenes for hyperparameter selection, and report all results on the validation split. Experiments begin with the car class.

\textit{Distance-binned metrics.} Because the expected benefit of radar is at long range, every metric is reported separately for objects at 0--30\,m, 30--50\,m, 50--100\,m, and 100--200\,m from the ego vehicle. For each bin we report 3D average precision (AP), recall, and center-distance AP. Center-distance AP matches predictions to ground truth by BEV center distance rather than IoU, following nuScenes~\cite{caesar2020nuscenes}, because small position errors at long range would otherwise dominate an IoU-based score. We also report the latency that the radar features and fusion network add per frame.

\textit{Baselines and ablations.} We compare against two baselines trained on aiMotive under the same split: the LiDAR-only 3D detector with its original scores, and camera-LiDAR CLOCs (set R0). Together they isolate the contributions of the camera and of radar. Each radar feature set in Table~\ref{tab:featsets} is then compared with R0 to measure which radar cues carry the gain and whether adding more features helps or introduces noise. To measure how the method behaves when the camera is unavailable or degraded, for example at night, the selected feature set is also evaluated with all 2D candidates removed at test time, so that every 3D candidate uses the reserved no-camera row. The LiDAR-radar and three-sensor baselines released with aiMotive~\cite{matuszka2023aimotive} serve as external reference points, although they use different detectors.

\subsection{Implementation Details}
The fusion network and training pipeline build on the official CLOCs implementation~\cite{pang2020clocs}, updated to run with PyTorch 1.10 and CUDA 11.3 on current GPUs. The fixed candidate dimensions of the original code are generalized so that the number of 3D candidates follows the extended aiMotive anchor grid. The 3D detector is trained with OpenPCDet~\cite{openpcdet2020}. Experiments run on NVIDIA A100 GPUs, and the latency added by computing the radar features and running the fusion network is measured for each feature set on a single GPU with batch size one, excluding the cost of the two detectors, which is reported separately.

\section{Experiments}\label{sec:exp}
This section reports the experiments that follow the protocol of Section~\ref{sec:method}, and Section~\ref{sec:exp-setup} states where a run departs from it. Fusion results are means over three training seeds. Every number comes from the result files and logs in our repository. Changes made after the validation split had been scored are marked as post hoc ($^{\dagger}$).

\begin{table*}[t]
\caption{Results on aiMotive Validation (Daytime Clear Weather, 4,292 Frames)}
\label{tab:main}
\centering
\scriptsize
\setlength{\tabcolsep}{4pt}
\begin{tabular}{@{}l cccccc cccccc@{}}
\toprule
& \multicolumn{6}{c}{\textbf{APBEV@0.3}} & \multicolumn{6}{c}{\textbf{Center-distance AP (APCD)}} \\
\cmidrule(lr){2-7} \cmidrule(l){8-13}
\textbf{Method} & 0--30 & 30--50 & 50--100 & 100--200 & $>$50 & All & 0--30 & 30--50 & 50--100 & 100--200 & $>$50 & All \\
\midrule
SECOND & 0.975 & 0.847 & 0.725 & 0.277 & 0.517 & 0.630 & 0.909 & 0.664 & 0.615 & 0.280 & 0.458 & 0.556 \\
CLOCs (R0) & 0.974 & 0.857 & 0.701 & 0.267 & 0.493 & 0.599 & 0.926 & 0.693 & 0.589 & 0.252 & 0.424 & 0.525 \\
CLOCs-R (R2) & 0.974 & 0.870 & 0.706 & 0.271 & 0.497 & 0.607 & 0.918 & 0.698 & 0.593 & 0.255 & 0.428 & 0.530 \\
R0 + residual$^{\dagger}$ & 0.974 & 0.861 & 0.723 & 0.296 & 0.521 & 0.623 & 0.925 & 0.697 & 0.606 & 0.280 & 0.448 & 0.545 \\
R2 + residual$^{\dagger}$ & 0.974 & 0.874 & 0.727 & 0.298 & 0.524 & 0.630 & 0.922 & 0.704 & 0.609 & 0.283 & 0.451 & 0.551 \\
\bottomrule
\end{tabular}

\vspace{4pt}
\setlength{\tabcolsep}{3pt}
\begin{tabular}{@{}l ccccc ccccc ccccc@{}}
\toprule
& \multicolumn{5}{c}{\textbf{AP3D@0.5}} & \multicolumn{5}{c}{\textbf{AP3D@0.7}} & \multicolumn{5}{c}{\textbf{Recall@BEV0.3}} \\
\cmidrule(lr){2-6} \cmidrule(lr){7-11} \cmidrule(l){12-16}
\textbf{Method} & 0--30 & 30--50 & 50--100 & 100--200 & All & 0--30 & 30--50 & 50--100 & 100--200 & All & 0--30 & 30--50 & 50--100 & 100--200 & All \\
\midrule
SECOND & 0.950 & 0.514 & 0.323 & 0.052 & 0.336 & 0.393 & 0.041 & 0.024 & 0.001 & 0.066 & 0.987 & 0.920 & 0.858 & 0.650 & 0.807 \\
CLOCs (R0) & 0.958 & 0.603 & 0.328 & 0.047 & 0.349 & 0.413 & 0.047 & 0.018 & 0.001 & 0.063 & 0.987 & 0.924 & 0.858 & 0.614 & 0.794 \\
CLOCs-R (R2) & 0.957 & 0.595 & 0.329 & 0.047 & 0.350 & 0.381 & 0.044 & 0.018 & 0.001 & 0.062 & 0.988 & 0.924 & 0.860 & 0.623 & 0.798 \\
R0 + residual$^{\dagger}$ & 0.958 & 0.602 & 0.335 & 0.050 & 0.354 & 0.416 & 0.047 & 0.020 & 0.001 & 0.066 & 0.988 & 0.925 & 0.871 & 0.655 & 0.815 \\
R2 + residual$^{\dagger}$ & 0.958 & 0.599 & 0.336 & 0.050 & 0.354 & 0.405 & 0.045 & 0.020 & 0.001 & 0.066 & 0.988 & 0.927 & 0.872 & 0.659 & 0.817 \\
\bottomrule
\end{tabular}

\vspace{2pt}
\parbox{\textwidth}{\scriptsize Distance bins in meters; validation has 17,669 cars (2,474, 1,470, 6,908 and 6,777 in the four bins), and $>$50 pools the last two bins, the selection metric. R0 is CLOCs without radar and R2 is CLOCs-R with the selected feature set. Fusion rows are means of 3 seeds. $^{\dagger}$Post hoc residual scoring, Eq.~\eqref{eq:residual}. Recall is limited by the 100-box cap (Section~\ref{sec:exp-setup}).}
\end{table*}

\subsection{Dataset}\label{sec:exp-data}
\textit{Source and subset.} We use the aiMotive dataset~\cite{matuszka2023aimotive}, restricted to its daytime clear-weather highway and urban scenes; night and rain scenes are left for future work. In each of the two conditions, every tenth official training scene is held out for selection, which leaves 109 training scenes and 13 holdout scenes. All 21 official validation scenes in these conditions are used for reporting. To limit storage, every second annotated frame of the training and holdout scenes is kept, giving 7,753 training, 923 holdout and 4,292 validation frames.

\textit{Labels and scored region.} The car class combines the CAR, VAN and PICKUP labels. Trucks, buses and trailers are ignored: candidates that overlap them (BEV IoU of at least 0.1) count as neither true nor false positives, in training or in evaluation. An object is scored when its center lies inside the SECOND grid ($0 \le x < 204.8$\,m, $|y| < 25.6$\,m) and inside the front camera's view. The validation split then contains 17,669 cars, and holdout contains 6,242, of which only 466 are at 100--200\,m.

\textit{Label shift.} The training scenes are auto-annotated with manual checks, while the validation scenes were labeled by hired annotators~\cite{matuszka2023aimotive}. The car labels differ accordingly. The median length is 4.00\,m in training, where 43\% of cars are exactly 4.00\,m long (58\% in holdout), and 4.37\,m in validation, where none are. Detectors learn the training convention, so strict IoU metrics on validation reflect label differences as well as detection errors (Section~\ref{sec:exp-det}).

\textit{Composition.} The 21 validation scenes come from five drives, and 83\% of the validation cars beyond 50\,m (11,352 of 13,685) come from nine consecutive urban scenes of one drive, about 2.25 minutes long. Five validation scenes and one holdout scene pass through tunnels, where the camera is badly underexposed, and no training scene does. Eleven of the 13 holdout scenes come from drives that also contribute training scenes, so holdout measures generalization to unseen scenes rather than unseen drives.

\textit{Preprocessing.} 2D training labels are generated as in Section~\ref{sec:method}. Returns from the front long-range radar are accumulated over $M = 3$ sweeps and compensated with Eq.~\eqref{eq:comp}. In 33 of the 12,876 validation sweeps (0.26\%), the ego poses have no timestamps, and the ego velocity is then computed with the sequence's frame period. We checked the compensation on all 143 sequences. In the median sequence, 88\%, 90\% and 91\% of the returns away from labeled objects are near static (below 1\,m/s) in training, holdout and validation, and Doppler speeds on labeled cars differ from their labeled velocities by a median of 0.06, 0.08 and 0.18\,m/s. No sequence fails both checks.

\subsection{Experimental Setup}\label{sec:exp-setup}
\textit{Hardware and software.} All runs use one 20\,GB NVIDIA A100 virtual GPU (A100X-20C) on the Jetstream2 cloud, with Python 3.8, PyTorch 1.10.2, CUDA 11.3 and spconv 1.2.1. SECOND is trained with OpenPCDet~\cite{openpcdet2020}, and Cascade R-CNN with MMDetection 2.28.2~\cite{chen2019mmdetection}.

\textit{SECOND.} The network follows the OpenPCDet SECOND configuration and is trained from random initialization for 40 epochs with batch size 4, using Adam with a one-cycle schedule (maximum learning rate $3{\times}10^{-3}$) and weight decay 0.01, which took about 2\,h 50\,min. The grid covers $[0, 204.8] \times [-25.6, 25.6] \times [-1.5, 2.5]$\,m with 0.1\,m voxels, and one car anchor of $4.2 \times 2.0 \times 1.55$\,m is placed at two yaw angles. Augmentation uses random left-right flips, rotations of up to $\pm 5^{\circ}$ (larger rotations move distant cars out of the narrow grid), and scaling by 0.95--1.05.

\textit{Cascade R-CNN.} A ResNet-50 FPN Cascade R-CNN pretrained on COCO~\cite{lin2014coco} is fine-tuned for 4 epochs with SGD (learning rate 0.005, batch size 4) at an input size of $1280 \times 704$, which took about 1\,h 10\,min.

\textit{Candidates.} Cascade R-CNN keeps its top $K_{2D} = 200$ boxes per image. SECOND produces 32,768 anchor-level boxes per frame, and storing all of them did not fit on our disk, so the top 2,000 by score are kept, rather than every anchor as Section~\ref{sec:method} states. These contain a box with BEV IoU of at least 0.3 for 85\% of the validation cars at 100--200\,m, and for 59\% of the holdout cars.

\textit{Fusion network.} The network, inputs and loss follow Section~\ref{sec:method}. It is trained for 3 epochs over the 7,753 training frames with a batch size of one frame, a one-cycle schedule (maximum learning rate $3{\times}10^{-3}$, 40\% warm-up) and weight decay 0.01. The optimizer is AdamW~\cite{loshchilov2019adamw}, the decoupled-weight-decay form of Adam. Training targets use BEV IoU and the reserved row uses $-1$, as Section~\ref{sec:method} states, whereas the official CLOCs code uses 3D IoU and $-10$. Each configuration is trained with seeds 1, 2 and 3, at about 2 minutes per seed. GPU training is not bit-identical between runs: retraining one seed reproduced its scores to within 0.005. The gate was selected on holdout with R3, from $\delta_0 \in \{0.25, 0.5, 1.0\}$\,m and $\kappa \in \{0.01, 0.02, 0.04\}$, giving $\delta_0 = 0.5$\,m and $\kappa = 0.02$, and $v_{\max} = 40$\,m/s.

\textit{Evaluation.} Detections are evaluated with no score threshold after BEV NMS at IoU 0.1, which keeps at most 1,000 boxes before and 100 after NMS per frame. Matching is greedy in score order, and AP is all-point interpolated. Each ground-truth box and detection falls in a distance bin by its own BEV range, and only pairs in the same bin can match. Besides 3D AP at IoU 0.5 and 0.7, recall, and center-distance AP (APCD, the mean AP at 0.5, 1, 2 and 4\,m~\cite{caesar2020nuscenes}), we report BEV AP at IoU 0.3 (APBEV), the matching threshold of the aiMotive baselines~\cite{matuszka2023aimotive}, because the label shift makes strict IoU thresholds unreliable on this dataset. Recall uses BEV IoU 0.3. The selection metric, APCD beyond 50\,m, pools the 50--100 and 100--200\,m bins. The 100-box limit is reached on 99\% of validation frames. Raising it to 500, with 2,000 boxes before NMS, changes APBEV and APCD by at most 0.003, but raises recall at 100--200\,m from 0.650 to 0.695 for SECOND and from 0.659 to 0.695 for CLOCs-R with residual scoring, so recall differences between methods should be read with care.

\subsection{KITTI Verification}\label{sec:exp-kitti}
With the authors' released weights, CLOCs reproduces the published car 3D AP of 92.75 / 81.28 / 77.70 (easy / moderate / hard) on KITTI validation in our environment. SECOND alone gives 90.82 / 79.73 / 76.87, against 90.97 / 79.94 / 77.09 in the CLOCs repository, and a two-epoch training run of the official code reaches 92.01 / 80.82 / 77.42. We then ran our own fusion implementation, the one used on aiMotive, on the authors' KITTI candidates (3 epochs, 3 seeds). It matches the authors' CLOCs to within 0.006 on every metric and distance bin (Table~\ref{tab:kitti}).

\begin{table}[t]
\caption{Our Fusion Implementation on KITTI Validation (3,769 Frames)}
\label{tab:kitti}
\centering
\footnotesize
\setlength{\tabcolsep}{3.5pt}
\begin{tabular}{@{}lcccccc@{}}
\toprule
& \multicolumn{3}{c}{\textbf{APBEV@0.3}} & \multicolumn{3}{c}{\textbf{AP3D@0.7}} \\
\cmidrule(lr){2-4} \cmidrule(l){5-7}
\textbf{Model} & All & 30--50 & 50--70 & All & 30--50 & 50--70 \\
\midrule
SECOND & 0.929 & 0.867 & 0.482 & 0.704 & 0.496 & 0.104 \\
Ours (3 seeds) & 0.942 & 0.906 & 0.621 & 0.719 & 0.542 & 0.160 \\
Authors' CLOCs & 0.941 & 0.906 & 0.622 & 0.717 & 0.536 & 0.156 \\
\bottomrule
\multicolumn{7}{@{}l}{Distance bins in meters; 50--70 ends at 70.4\,m.}
\end{tabular}
\end{table}

\subsection{Detector Baselines}\label{sec:exp-det}
Table~\ref{tab:detectors} compares SECOND on aiMotive with SECOND on KITTI. Within KITTI's 70.4\,m range, aiMotive is level with KITTI at lenient matching: APBEV is 0.975 against 0.973 at 0--30\,m, 0.847 against 0.867 at 30--50\,m, and 0.738 against 0.481 at 50--70.4\,m. Its lower score over 0--70.4\,m (0.867 against 0.929) comes from the distance mix, since 47\% of aiMotive cars in that range are at 50--70.4\,m, against 9\% on KITTI. The full-range score falls to 0.630 because 38\% of all validation cars are at 100--200\,m, where APBEV is 0.277. At strict matching the gap is large (AP3D@0.7 of 0.066 against 0.704), and much of it is the label shift: matched validation boxes are about 0.2\,m too wide at up to 100\,m, and the median BEV IoU is about 0.7, so APBEV falls from 0.699 at IoU 0.1 to 0.207 at IoU 0.7.

\begin{table}[t]
\caption{SECOND Alone on aiMotive and KITTI Validation}
\label{tab:detectors}
\centering
\footnotesize
\setlength{\tabcolsep}{3.5pt}
\begin{tabular}{@{}lcccccc@{}}
\toprule
& \multicolumn{3}{c}{\textbf{aiMotive}} & \multicolumn{3}{c}{\textbf{KITTI}} \\
\cmidrule(lr){2-4} \cmidrule(l){5-7}
\textbf{Distance} & BEV@0.3 & APCD & 3D@0.7 & BEV@0.3 & APCD & 3D@0.7 \\
\midrule
0--30\,m & 0.975 & 0.909 & 0.393 & 0.973 & 0.971 & 0.854 \\
30--50\,m & 0.847 & 0.664 & 0.041 & 0.867 & 0.861 & 0.496 \\
50--70.4\,m & 0.738 & 0.628 & 0.044 & 0.481 & 0.467 & 0.104 \\
0--70.4\,m & 0.867 & 0.748 & 0.158 & 0.929 & 0.925 & 0.704 \\
50--100\,m & 0.725 & 0.615 & 0.024 & -- & -- & -- \\
100--200\,m & 0.277 & 0.280 & 0.001 & -- & -- & -- \\
All & 0.630 & 0.556 & 0.066 & 0.929 & 0.925 & 0.704 \\
\bottomrule
\multicolumn{7}{@{}l}{KITTI's SECOND grid ends at 70.4\,m.}
\end{tabular}
\end{table}

At long range, LiDAR coverage is the main limit. On holdout, 49\% of cars at 100--200\,m have no LiDAR points and 66\% have fewer than five, and counting only cars with at least five points raises APBEV there from 0.313 to 0.817 (158 cars). Both detectors also score much higher on frames they were trained on: on 900 training frames, SECOND reaches APBEV 0.835 and Cascade R-CNN AP2D@0.5 0.803, against 0.629 and 0.499 on holdout. Cascade R-CNN reaches AP2D@0.5 of 0.431 on validation against 0.827 on KITTI, but 0.827 against 0.812 for boxes at least 40 pixels tall, and 57\% of aiMotive cars are under 25 pixels tall, against 12\% on KITTI. At 100--200\,m, its candidates overlap only 32\% of the validation cars at IoU 0.5.

\subsection{Main Results}\label{sec:exp-main}
Table~\ref{tab:main} compares the fusion models with SECOND alone. Camera-LiDAR CLOCs (R0) improves on SECOND up to 50\,m, most clearly in APCD (0.664 to 0.693 at 30--50\,m) and in AP3D@0.5 at 30--50\,m (0.514 to 0.603). Beyond 50\,m it loses: APCD there falls from 0.458 to 0.424, and full-range APBEV from 0.630 to 0.599. On KITTI the same code gains most at long range, and on aiMotive the loss also appears within KITTI's range (APBEV at 50--70.4\,m falls from 0.738 to 0.704). With the selected feature set R2, radar raises APBEV at 30--50\,m from 0.857 to 0.870 but leaves the selection metric almost unchanged (0.428 against 0.424), and both fusion models stay below SECOND beyond 50\,m.

\textit{Residual scoring (post hoc).} At 100--200\,m, CLOCs ranks SECOND's boxes worse than SECOND's own score does (AUC 0.873 against 0.909 for separating real cars from false detections). After this analysis, and therefore after the validation split had been scored, we tested six changes on holdout under a rule fixed in advance (Section~\ref{sec:exp-abl}). Only one was adopted: the fusion network corrects SECOND's score instead of replacing it,
\begin{equation}
\hat{s}_j = \operatorname{logit}\big(s^{3D}_j\big) + \max_{i} f_{\theta}(\mathbf{T}_{ij}),
\label{eq:residual}
\end{equation}
where the last layer of $f_{\theta}$ is initialized to zero, so that training starts from SECOND's ranking. This adds no parameters. Residual scoring recovers most of the loss beyond 50\,m, where APCD rises to 0.448 without radar and to 0.451 with R2. APBEV beyond 50\,m (0.524) and over the full range (0.630) then match or exceed SECOND. On top of residual scoring, radar adds little: 0.003 in APCD beyond 50\,m, and 0.013 in APBEV at 30--50\,m (0.861 to 0.874). SECOND keeps a lead in APCD beyond 50\,m (0.458 against 0.451) and at 50--100\,m, but Section~\ref{sec:exp-analysis} shows that this lead is not significant.

\subsection{Comparison with the aiMotive Baselines}\label{sec:exp-prior}
Table~\ref{tab:prior} places our models next to the baselines released with aiMotive~\cite{matuszka2023aimotive}. These use different networks (a VoxelNet LiDAR branch and BEVFusion-style fusion), evaluate four object categories class-agnostically with Hungarian matching, and score a region that extends behind the vehicle, while our models score cars only, inside the front camera's view, with greedy matching. The numbers therefore indicate the expected range rather than a ranking. Our SECOND scores below their LiDAR baseline on highways (0.639 against 0.755), slightly above it in urban scenes (0.633 against 0.608), and below it beyond 75\,m in both (0.426 and 0.405 against 0.526 and 0.470). Their radar fusion raises AP beyond 75\,m in urban scenes (0.470 to 0.525) but not on highways (0.526 to 0.522), whereas radar raises ours by at most 0.006 beyond 75\,m in either condition. Their camera fusion raises full-range AP in both conditions, while camera-LiDAR CLOCs lowers ours until residual scoring is added.

\begin{table}[t]
\caption{Reference Comparison with the aiMotive Baselines~\cite{matuszka2023aimotive} (All-Point BEV AP@0.3, Validation)}
\label{tab:prior}
\centering
\footnotesize
\setlength{\tabcolsep}{4pt}
\begin{tabular}{@{}llcccc@{}}
\toprule
& & \multicolumn{2}{c}{\textbf{Full range}} & \multicolumn{2}{c}{\textbf{$>$75\,m}} \\
\cmidrule(lr){3-4} \cmidrule(l){5-6}
\textbf{Model} & \textbf{Sensors} & Highway & Urban & Highway & Urban \\
\midrule
aiMotive & L & 0.755 & 0.608 & 0.526 & 0.470 \\
aiMotive & L+C & 0.772 & 0.656 & 0.509 & 0.466 \\
aiMotive & L+R & 0.769 & 0.617 & 0.522 & 0.525 \\
aiMotive & L+R+C & 0.762 & 0.644 & -- & -- \\
\midrule
SECOND & L & 0.639 & 0.633 & 0.426 & 0.405 \\
R0 & L+C & 0.594 & 0.603 & 0.379 & 0.399 \\
R2 & L+R+C & 0.599 & 0.613 & 0.385 & 0.403 \\
R0 + res.$^{\dagger}$ & L+C & 0.635 & 0.623 & 0.435 & 0.418 \\
R2 + res.$^{\dagger}$ & L+R+C & 0.639 & 0.632 & 0.441 & 0.420 \\
\bottomrule
\multicolumn{6}{@{}L{0.97\columnwidth}}{The baselines' $>$75\,m values come from their longer-trained models, and no three-sensor value is reported there. Our rows (SECOND, R0 = CLOCs, R2 = CLOCs-R, res.\ = residual scoring) score cars only in daytime clear-weather scenes and are means of 3 seeds; our urban rows cover 14,625 cars, mostly from one drive.}
\end{tabular}
\end{table}

\subsection{Ablation}\label{sec:exp-abl}
\textit{Radar feature sets.} Table~\ref{tab:ablation} reports every set of Table~\ref{tab:featsets}. Beyond 50\,m, all eight sets lie within 0.007 of each other on validation and within 0.006 on holdout, about the spread between seeds of one set (validation standard deviation up to 0.010). R2 (Doppler speed alone) was selected under the rule of Section~\ref{sec:radarfeat}, with 0.4554 on holdout, but R7 (0.4550) and R3 (0.4547) are effectively tied with it. A retraining of R2 during tuning scored 0.4545, which would have ranked it below both, so the selection is not stable and does not show that Doppler speed is the best single cue. More features do not help: R7, with all five, is no better than R2 on either split. The consistent radar effect is at 30--50\,m, where every set raises validation APBEV from 0.857 to between 0.866 and 0.871, and support alone (R1) already reaches 0.866. At strict close-range matching, the sets with Doppler speed lose AP3D@0.7 at 0--30\,m on validation (0.375--0.386 against 0.413 for R0). This does not repeat on holdout, however, where R5, which has no Doppler speed, loses the most (0.465 against 0.492).

\begin{table}[t]
\caption{Radar Feature Sets (Table~\ref{tab:featsets}): Accuracy and Added Latency}
\label{tab:ablation}
\centering
\footnotesize
\setlength{\tabcolsep}{3.5pt}
\begin{tabular}{@{}lccccc@{}}
\toprule
& \multicolumn{2}{c}{\textbf{APCD $>$50\,m}} & \textbf{BEV@0.3} & \textbf{3D@0.7} & \textbf{Latency} \\
\cmidrule(lr){2-3}
\textbf{Set} & Holdout & Val. & 30--50\,m & 0--30\,m & (ms) \\
\midrule
SECOND & 0.4601 & 0.458 & 0.847 & 0.393 & -- \\
R0 & 0.4504 & 0.424 & 0.857 & 0.413 & 1.5 \\
R1 & 0.4509 & 0.424 & 0.866 & 0.417 & 4.8 \\
R2 (selected) & 0.4554 & 0.428 & 0.870 & 0.381 & 19.0 \\
R3 & 0.4547 & 0.426 & 0.867 & 0.384 & 19.2 \\
R4 & 0.4542 & 0.427 & 0.868 & 0.375 & 44.8 \\
R5 & 0.4495 & 0.421 & 0.866 & 0.411 & 21.5 \\
R6 & 0.4536 & 0.422 & 0.870 & 0.386 & 32.0 \\
R7 & 0.4550 & 0.423 & 0.871 & 0.384 & 62.1 \\
\bottomrule
\multicolumn{6}{@{}L{0.97\columnwidth}}{Means of 3 seeds. Holdout chose the set, and validation played no part. The BEV@0.3 and 3D@0.7 columns are on validation. Latency is the median per frame of the radar features (one CPU thread) and the fusion forward pass (GPU), excluding the detectors.}
\end{tabular}
\end{table}

\textit{Camera removed.} With all 2D candidates removed at test time, APCD beyond 50\,m on validation rises for both R0 (0.424 to 0.444) and R2 (0.428 to 0.448). On holdout, it rises for R0 (0.450 to 0.460) but falls for R2 (0.455 to 0.439). The reserved row is almost untrained, because only 78 of about 50,000 confident training candidates lack an overlapping 2D box. The test therefore describes the camera's effect on these models at long range, not a model trained to work without a camera, and it does not cover night driving, which our subset excludes.

\textit{Gate.} Across the nine gates, holdout APCD beyond 50\,m spans 0.4522 to 0.4551, about the spread between the seeds of one gate (0.452--0.459). Every gate beats R0 (0.450), and none reaches SECOND (0.460), so the results are not sensitive to the gate within this grid.

\textit{Fusion changes (post hoc).} Table~\ref{tab:tuning} lists the six changes tried after the as-designed results. Each was tested alone against R2 on holdout and adopted only if its mean beat R2's by more than the standard deviation of R2's three seeds, a bar of 0.4582 fixed before the runs. The winner was scored once on validation and also trained without radar. Only residual scoring passed. Far-range targets lowered the score, and neither more radar sweeps with motion correction nor a radar cross-section channel raised it.

\begin{table}[t]
\caption{Post Hoc Fusion Changes, Selected on Holdout (APCD $>$50\,m, Mean of 3 Seeds)}
\label{tab:tuning}
\centering
\footnotesize
\setlength{\tabcolsep}{4pt}
\begin{tabular}{@{}L{0.62\columnwidth}cc@{}}
\toprule
\textbf{Change to R2} & \textbf{Holdout} & \textbf{Adopted} \\
\midrule
None (R2 retrained) & 0.4545 & -- \\
Far-range targets: positive at BEV IoU $\ge$0.5, negative $\le$0.3 beyond 50\,m & 0.4433 & No \\
Residual scoring, Eq.~\eqref{eq:residual} & \textbf{0.4619} & \textbf{Yes} \\
SECOND's score as a scaled logit input & 0.4502 & No \\
$M = 5$ sweeps with motion correction & 0.4552 & No \\
$M = 10$ sweeps with motion correction & 0.4550 & No \\
Median radar cross-section as an extra channel & 0.4529 & No \\
\bottomrule
\multicolumn{3}{@{}L{0.97\columnwidth}}{Adoption bar 0.4582; SECOND scores 0.4601 on holdout.}
\end{tabular}
\end{table}

\textit{Runtime.} Radar features dominate the added latency (Table~\ref{tab:ablation}), with medians of 3.7 to 61.0\,ms per frame in unoptimized single-threaded Python, against 1.1 to 1.5\,ms for the fusion forward pass. With batch size one on 200 validation frames, the detectors themselves take a median of 20.9\,ms (SECOND, from GPU transfer to the top-2,000 candidates, with voxelization done by the data loader) and 50.6\,ms (Cascade R-CNN at $1280 \times 704$, including preprocessing but not image decoding) per frame, so the 19.0\,ms that R2 adds compares with 71.5\,ms for the two detectors run one after the other.

\subsection{Analysis}\label{sec:exp-analysis}
\textit{Uncertainty.} Table~\ref{tab:boot} gives 95\% intervals from resampling scenes with replacement (1,000 resamples), with the trained models held fixed and their three seeds averaged. CLOCs' loss to SECOND beyond 50\,m and the gain from residual scoring are clear on validation. CLOCs-R with residual scoring and SECOND are statistically tied. Radar's effect beyond 50\,m is small but positive on both splits. It is, however, smaller than the variation between training seeds (standard deviation about 0.004), and the intervals treat scenes as independent although many share a drive, so they are likely too narrow.

\begin{table}[t]
\caption{Differences with 95\% Scene-Bootstrap Intervals (APCD $>$50\,m Unless Stated)}
\label{tab:boot}
\centering
\scriptsize
\setlength{\tabcolsep}{2.5pt}
\begin{tabular}{@{}L{0.33\columnwidth}cc@{}}
\toprule
\textbf{Comparison} & \textbf{Validation} & \textbf{Holdout} \\
\midrule
R0 $-$ SECOND & $-.034$ [$-.057$, $-.011$] & $-.010$ [$-.023$, $+.008$] \\
R0 + residual $-$ R0 & $+.024$ [$+.018$, $+.033$] & $+.008$ [$+.004$, $+.010$] \\
R2 + residual $-$ R0 + residual (radar) & $+.003$ [$+.000$, $+.005$] & $+.004$ [$+.002$, $+.006$] \\
R2 + residual $-$ SECOND & $-.007$ [$-.026$, $+.012$] & $+.002$ [$-.010$, $+.018$] \\
Radar with residual, APBEV@0.3, 30--50\,m & $+.014$ [$-.002$, $+.028$] & $+.009$ [$+.002$, $+.017$] \\
R2 + residual $-$ SECOND, APCD, 30--50\,m & $+.040$ [$+.016$, $+.063$] & $+.010$ [$-.005$, $+.022$] \\
\bottomrule
\end{tabular}
\end{table}

\textit{Why radar adds little at range.} At 100--200\,m on validation, 62\% of the real cars among SECOND's detections have a radar return in their gate, but so do 39\% of the false detections, and the radar features alone separate the two only weakly (AUC 0.615 for support and 0.664 for Doppler speed, against 0.909 for SECOND's score). A logistic model of SECOND's score with and without radar, scored on scenes it was not fit on, gains at most 0.008 AUC there. The trained R3 networks also barely use radar at that range: replacing the radar inputs with ``no return'' changes the fused logit of real cars by about $+0.5$ at 30--50\,m but by $-0.02$ at 100--200\,m. Training offers little signal there, since only 7,232 of 136,332 training positives (5\%) are at 100--200\,m, and SECOND looks nearly perfect on its own training frames at that range (AUC 0.984, against 0.909 on validation).

\textit{Failure modes.} Four effects limit the results. First, most of SECOND's remaining lead comes from the tunnel scenes, where the camera is dark. Beyond 50\,m there, APCD is 0.461 for SECOND against 0.431 for CLOCs-R with residual scoring (and 0.346 for R0), while without the tunnel scenes the two score 0.458 and 0.456. Second, camera evidence is weakest where fusion was meant to help: only 8\% of real cars at 100--200\,m on validation have a well-aligned 2D box (2D-3D IoU of at least 0.5) scoring above 0.5, against 97\% at 0--30\,m. Third, rescoring cannot recover cars that SECOND never proposed: among the 2,000 kept candidates, 15\% of the validation cars at 100--200\,m and 41\% of those on holdout have none with BEV IoU of at least 0.3. Fourth, strict 3D AP is bounded by the label shift, which rescoring cannot correct because box geometry comes from SECOND.

\textit{Disclosure.} The detectors were frozen, and the gate and the feature set were selected, using holdout only. Residual scoring was proposed after an analysis of validation results, however, so validation is not an untouched test set for it, and a clean estimate would need new data. For this reason, Table~\ref{tab:main} reports the as-designed and post hoc models side by side.

\section*{AI Usage Statement}
Claude Opus 5.5 was used to format our original draft made in Google Docs to conform to the IEEE LaTeX template provided on Canvas. It was also used to generate LaTeX code for the comparison table and verify that this document followed the requirements outlined in Canvas for this phase of the project.

\bibliographystyle{IEEEtran}
\bibliography{refs}

\end{document}